\documentclass[conference]{IEEEtran}
\IEEEoverridecommandlockouts
\usepackage{cite}
\usepackage{amsmath,amssymb,amsfonts}
\usepackage{algorithmic}
\usepackage{graphicx}
\usepackage{textcomp}
\usepackage{xcolor, soul}
\usepackage{lipsum}
\usepackage{algorithm}
\usepackage{multirow}
\usepackage{hyperref}
\usepackage{booktabs}

\renewcommand{\thefootnote}{}

\def\BibTeX{{\rm B\kern-.05em{\sc i\kern-.025em b}\kern-.08em
    T\kern-.1667em\lower.7ex\hbox{E}\kern-.125emX}}
\begin{document}

\makeatletter
\newcommand\blfootnote[1]{%
  \begingroup
  \renewcommand\thefootnote{}\footnote{#1}%
  \addtocounter{footnote}{-1}%
  \endgroup
}
\newcommand{\linebreakand}{%
  \end{@IEEEauthorhalign}
  \hfill\mbox{}\par
  \mbox{}\hfill\begin{@IEEEauthorhalign}
}
\makeatother

\title{Generalizable Reinforcement Learning with Biologically Inspired Hyperdimensional Occupancy Grid Maps for Exploration and Goal-Directed Path Planning}

\author{\IEEEauthorblockN{Shay Snyder*\thanks{* These authors contributed equally to this work.}}
\IEEEauthorblockA{
\textit{George Mason University}\\ssnyde9@gmu.edu}
\and
\IEEEauthorblockN{Ryan Shea*}
\IEEEauthorblockA{\textit{Columbia University}\\rs4235@columbia.edu}
\and
\IEEEauthorblockN{Andrew Capodieci}
\IEEEauthorblockA{\textit{Neya Robotics}\\acapodieci@neyarobotics.com}
\linebreakand
\IEEEauthorblockN{David Gorsich}
\IEEEauthorblockA{\textit{US Army Futures Command}\\david.j.gorsich.civ@army.mil}
\and
\IEEEauthorblockN{Maryam Parsa$^\dagger$\thanks{$\dagger$ Corresponding author}}
\IEEEauthorblockA{\textit{George Mason University}\\mparsa@gmu.edu}
}

\maketitle

\begin{abstract}
Real-time autonomous systems utilize multi-layer computational frameworks to perform critical tasks such as perception, goal finding, and path planning. Traditional methods implement perception using occupancy grid mapping (OGM), segmenting the environment into discretized cells with probabilistic information. This classical approach is well-established and provides a structured input for downstream processes like goal finding and path planning algorithms. Recent approaches leverage a biologically inspired mathematical framework known as vector symbolic architectures (VSA), commonly known as hyperdimensional computing, to perform probabilistic OGM in hyperdimensional space. This approach, VSA-OGM, provides native compatibility with spiking neural networks, positioning VSA-OGM as a potential neuromorphic alternative to conventional OGM. However, for large-scale integration, it is essential to assess the performance implications of VSA-OGM on downstream tasks compared to established OGM methods. This study examines the efficacy of VSA-OGM against a traditional OGM approach, Bayesian Hilbert Maps (BHM), within reinforcement learning based goal finding and path planning frameworks, across a controlled exploration environment and an autonomous driving scenario inspired by the F1-Tenth challenge. Our results demonstrate that VSA-OGM maintains comparable learning performance across single and multi-scenario training configurations while improving performance on unseen environments by approximately 47\%. These findings highlight the increased generalizability of policy networks trained with VSA-OGM over BHM, reinforcing its potential for real-world deployment in diverse environments.
\end{abstract}


\section{Introduction}
\label{sec:introduction}


\noindent
Contemporary robotic systems are structured around a multi-level autonomy framework, where complex tasks are divided into interconnected sub-units. Each sub-unit focuses on specific functions, such as mapping, localization, goal identification, or path planning~\cite{robotics13010012}. Together, these sub-systems enable high-level capabilities, such as autonomous driving~\cite{REDA2024104630} and unknown environment exploration~\cite{zhao2024exploration}.

Localization and mapping have long been central to the multi-level autonomy framework, with simultaneous localization and mapping (SLAM) being a key technique for enabling autonomous agents to navigate complex environments~\cite{macario2022comprehensive}. SLAM provides agents with an information-rich representation of their surroundings, which can be leveraged by downstream goal-finding and path planning modules.
A traditional approach to the mapping component is Occupancy Grid Mapping (OGM)~\cite{elfes1989using}. These algorithms construct probabilistic representations of the environment, where each region is assigned a real value quantifying the likelihood of occupancy.

OGM literature is well-established, providing a structured framework for downstream processes such as goal finding and path planning algorithms at the expense of exponential Big O complexity~\cite{wilson2022convolutional}. Recently, biologically inspired mathematical frameworks, specifically vector symbolic architectures (VSA), have been applied to probabilistic OGM in hyperdimensional space~\cite{snyder2024brain}. This novel approach, known as Vector Symbolic Architectures for Occupancy Grid Mapping (VSA-OGM)~\cite{snyder2024brain}, supports integration with spiking neural networks, making it a promising neuromorphic alternative to conventional OGM techniques.
However, to facilitate large-scale integration with existing systems, it is crucial to evaluate VSA-OGM's performance on downstream tasks relative to traditional methods.

\footnotetext{DISTRIBUTION STATEMENT A. Approved for public release; distribution is unlimited. OPSEC \# 9329 approved for Release.}
This study investigates the efficacy of VSA-OGM compared to a well-established OGM technique, Bayesian Hilbert Maps (BHM)~\cite{senanayake2017bayesian}, within reinforcement learning (RL) frameworks for goal finding and path planning. Experiments are conducted across controlled environments for unknown environment exploration~\cite{Koutras2021MarsExplorer} and autonomous driving inspired by the F1-Tenth challenge~\cite{Brunnbauer_racecar_gym}. Our results indicate that VSA-OGM not only maintains robust learning performance but also significantly enhances policy generalizability to different starting conditions and diverse unseen environments.
In summary, the major contributions of this paper are as follows:
\begin{itemize}
    \item We design environment wrappers for two distinct reinforcement learning scenarios: one simulating a rover exploring unknown regions~\cite{Koutras2021MarsExplorer}, and another inspired by the F1-Tenth driving challenge~\cite{Brunnbauer_racecar_gym}. These wrappers enable seamless integration with OGM methods.
    \item Our results, conducted using various parameter combinations and environmental configurations to evaluate performance across a diverse range of scenarios, highlight that VSA-OGM achieves comparable performance to traditional approaches while improving policy generalization on unseen environments and different starting conditions by approximately 47\%.
    \item We perform multi-map training and evaluation across both environments with both OGM methods and VSA-OGM presents increased generalization, on average, across all evaluation maps and unseen scenarios.
\end{itemize}
\section{Background}
\label{sec:background}

\noindent
OGM techniques have been developed to incorporate semantically labeled spatial information and give autonomous agents critical information to accomplish diverse tasks such as unknown environment exploration~\cite{robotics13010012} and path planning~\cite{zhao2024exploration}. The field has diverged into three primary sub-fields with each being defined by unique computational attributes.

\textit{Traditional methods} are the most studied from the foundational work of Elfes~\cite{elfes1989using} to more recent approaches like BHM~\cite{senanayake2017bayesian}. These approaches perform dense probabilistic calculations to learn from semantically-labeled spatial information at the cost of cubic computational complexity. BHM is a flexible framework that allows users to dictate the retention of the entire covariance matrix or only the variance. This enables BHM to scale between offline mapping scenarios with cubic complexity or real-time edge applications sub-cubic complexity, respectively.  
A newer version of BHM, known as Fast Bayesian Hilbert Maps (Fast-BHM)~\cite{zhi2019continuous}, reduces computational complexity to sub-quadratic levels by assuming independence between voxels and not utilizing a full covariance matrix. We were unable to include Fast-BHM in this study because their code is not publically available.

With the advent in AI, \textit{neural methods} have emerged with the goal of compressing probabilistic information into the black-box latent space of weights and biases~\cite{evilog}. While these approaches are effective for OGM, the black-box nature of deep neural networks makes them difficult to validate in safety critical environments such as autonomous driving~\cite{REDA2024104630}. Moreover, these methods must be trained on each operating domain to avoid performance degradation from domain shift. We limit the scope of our comparison to only traditional methods because we are evaluating the generalizability of OGMs in unknown environments which would require domain specific retraining for \textit{neural methods}.

A third subfield, \textit{neuro-symbolic methods} have emerged. These approaches, Convolutional Bayesian Kernel Inference (ConvBKI)~\cite{wilson2022convolutional} and VSA-OGM~\cite{snyder2024brain} combine benefits of both prior methods. Although ConvBKI and VSA-OGM have shown improvements in algorithmic complexity with respect to environment size and density, their efficacy for downstream path planning and environmental exploration remains a critical concern because they have not been compared against traditional OGM methods.

VSA-OGM leverages VSAs, a framework also known as hyperdimensional computing. VSAs provide a way to represent and manipulate information using high-dimensional vectors, drawing parallels to how the brain encodes and processes data. This biologically motivated approach serves as a mathematical framework that approximates higher-order cognitive functions in neural networks~\cite{10.1016/0004-3702(90)90007-M, eliasmith2013build}. A specific VSA architecture, known as Spatial Semantic Pointers (SSPs)~\cite{eliasmith2013build}, enables probabilistic inference over continuous representations in hyperdimensional space through a process known as fractional binding~\cite{komer2020biologically}.

SSPs operate on high dimensional vectors that are unit length and generated by performing the inverse discrete Fourier transform on uniformly distributed phasors between $-\pi$ and $\pi$. SSPs utilize two primary operations: binding 
and bundling. Binding, analogous to multiplication, combines two or more input vectors into a final invertible
representation that is distinct from each input. Bundling, analogous to addition, combines multiple input
vectors into their superposition. SSPs follow the same vector operations described in Holographic Reduced Representations (HRR)~\cite{plate1991holographic} where binding is implemented as circular convolution and bundling is implemented as element-wise addition. SSPs support encoding continuous values by exponentiating an axis vector $\phi_x$ in the complex domain by the desired axis values $x$ through a process known as fractional binding:

\begin{equation}
    \phi(x) = \mathcal{F}^{-1}(\mathcal{F}(\phi)^{x/l}),
\end{equation}

where $l$ is the length-scale parameter~\cite{komer2020biologically}. Probabilistic inference is performed with SSPs through Hadamard products between vectors. This process returns a quasi-kernel density estimator over the axis approaching a sinc function as the vector dimensionality approaches infinity~\cite{furlong2022fractional}. The length scale parameter in the fractional binding operation controls the width of this kernel and can be used to adjust the resolution and noise resilience of the SSPs.

SSPs are the basis of VSA-OGM and enable probabilistic modelling of occupancy in hyperdimensional space. More information on the underlying mathematics of VSAs and HRRs can be found in~\cite{10.1016/0004-3702(90)90007-M, plate1991holographic, plate1994distributed, gayler2004vectorsymbolicarchitecturesanswer, eliasmith2013build}. More specific details on the extension of SSPs to create VSA-OGM can be found in~\cite{snyder2024brain}. Given this drastically different approach to probabilistic computation compared to traditional methods, a major question remains around the ability of downstream path planning algorithms to leverage the quasi-probabilistic properties of VSA-OGM. This stands as the central question of this work where we compare the efficacy of VSA-OGM against BHM for reinforcement learning based path planning and environmental exploration with diverse environments, scenarios, and parameter combinations.
\section{Reinforcement Learning with OGMs}
\label{sec:methods}


\noindent
Most OGM frameworks, such as BHM~\cite{senanayake2017bayesian} and VSA-OGM~\cite{snyder2024brain}, require a different data format versus the native LiDAR output of many reinforcement learning environments. OGM methods require formalized training data similar to the feature and target format defined in Scikit-Learn~\cite{pedregosa2011scikit}, where $X$ represents the data feature matrix with shape $\gamma_n \times 2$, where $\gamma_n$ is equal to the number of ray-casts in a LiDAR point cloud $\gamma$ and other 2 dimensions represent a Cartesian coordinate. Likewise, $y$ is a $\gamma_n\times1$ vector representing the labels of each point where $0$ signifies empty and $1$ signifies occupied.
Bridging this gap and extracting OGM training data from LiDAR-based RL environments requires the end user to augment baseline environments with a wrapping layer transforming environmental information into an OGM compatible format. In this section, we describe the RL environments utilized throughout all experiments and the environmental wrappers augmenting their baseline input and output structure into an OGM-friendly format.

\begingroup
\renewcommand{\arraystretch}{1.2}
\begin{table}[h]
\begin{center}
\caption{The mathematical symbols used throughout this paper.} 
\label{tab:symbol-table}  
\begin{tabular}{c|c}
\textbf{Symbol}   & \textbf{Meaning}                         \\ \hline
$\mathcal{G}$     & square grid of cells                     \\
$g$               & an individual grid cell in $\mathcal{G}$ \\
$g_{loc}$         & the agent's location in $\mathcal{G}$    \\
$N$               & the number of cells in $\mathcal{G}$     \\
$M$               & the number of ray-cast interpolations    \\
$\mathcal{M}$     & ground truth map of $\mathcal{G}$ cells  \\
$\gamma$          & a point cloud in polar coordinates       \\
$\gamma_n$        & the number of ray-casts in $\gamma$      \\
$\gamma_{max}$    & maximum cast distance in meters          \\
$\delta$          & normalized steering angle                \\
$\tau$            & normalized motor speed                   \\
$t$               & time                                     \\
$r$               & reward value                             \\
$r_{bonus}$       & reward for target map exploration        \\
$r_{explore}$     & reward for exploring new areas           \\
$r_{invalid}$     & penalty for invalid movement             \\
$r_{move}$        & penalty for movement                     \\
$r_{obstacle}$    & penalty for hitting an obstacle          \\
$r_{steer}$       & penalty for steering magnitude           \\
$r_{velocity}$    & reward for maintaining target velocity   \\
$r_{collision}$   & penalty for collisions                   \\
$v$               & velocity in meters per second            \\
$X$               & Cartesian ray-casts from $\gamma$        \\
$X_{int}$         & interpolated ray-casts from $\gamma$     \\
$y$               & occupancy label vector for $X$           \\
$y_{int}$         & occupancy label vector for $X_{int}$     \\
$\theta$          & angle in radians                         \\
$\Theta$          & angle vector for each ray-cast           \\
$\Delta\Theta$    & step-size between individual ray-casts   \\
$\lambda_x$       & reward scaling parameter                 \\
$T$               & orientation matrix                       \\

\end{tabular}
\end{center}
\end{table}
\endgroup

\textbf{\textit{Mars Explorer (MarsExplorer)}}~\cite{Koutras2021MarsExplorer}: \textit{MarsExplorer}~\cite{Koutras2021MarsExplorer} is an open-source Open-AI Gym~\cite{gym} compatible environment designed to train agents to explore randomized unknown environments. These environments are constrained to be a square grid $\mathcal{G}$ of $N=rows\times cols$ cells defined as:
\begin{equation}
    \mathcal{G} = \{(x, y): x\in[1,rows],y\in[1, cols] \}.
\end{equation}
The ground truth map of a scenario $\mathcal{M}$ is defined as a mapping from each individual grid cell $\mathcal{M}(g)$ such that:
\begin{equation}
    \mathcal{M}(g) = 
        \begin{cases} 
        0.3 & \text{empty} \\
        1 & \text{occupied} \\
        \end{cases} \hspace{8pt} g = (x, y) \in \mathcal{G}.
\end{equation}

The default observation space in \textit{MarsExplorer} is a real-valued matrix, with shape $\mathcal{G}$, indicating whether each cell is occupied or empty. The field of view is determined by propagating circular ray-casts uniformly distributed across the unit circle. These ray-casts simulate a LIDAR sensor, generating a point cloud vector $\gamma$ in polar coordinates with shape $\gamma_n$, where $\gamma_n$ represents the number of ray-casts in the point cloud. We utilized the default $\gamma_n=32$ provided within \textit{MarsExplorer}. The observation matrix is then updated by marking all intersected cells as empty ($g=0.3$) and terminal cells (those where a ray-cast terminates before the max cast distance) as occupied ($g=1.0$). Lastly, the agent's position $loc$ is marked by setting the corresponding cell to an intermediate value where $g_{loc}=0.6$. The action space for \textit{MarsExplorer} is discrete with 4 options representing moving one cell up, down, left, or right.

The reward $r$ is defined as a piecewise function consisting of four parts: $r_{explore}$, $r_{move}$, $r_{invalid}$, and $r_{bonus}$. With agents having the ultimate goal of exploring the entire environment, $r_{explore}$ is defined as the number of newly explored cells from $t-1$ to $t$. Therefore, we can also conclude that the entire grid has been explored when $\sum_{k=0}^Tr_{explore}(k) \rightarrow N$. To encourage the agent to use fewer movements to observe the environment, a fixed penalty $r_{move}$ of 0.5 is applied at every time step. $r_{bonus}$ serves as an encouragement bonus where the agent receives an increased reward of 100 if 95\% or more of the environment has been explored. Lastly, is $r_{invalid}$ which overrides the other pieces if activated. $r_{invalid}$ represents a fixed penalty of -100 if the agent hits an obstacle or tries to move out of bounds.

\begin{figure*}
    \centering
    \includegraphics[width=0.9\linewidth]{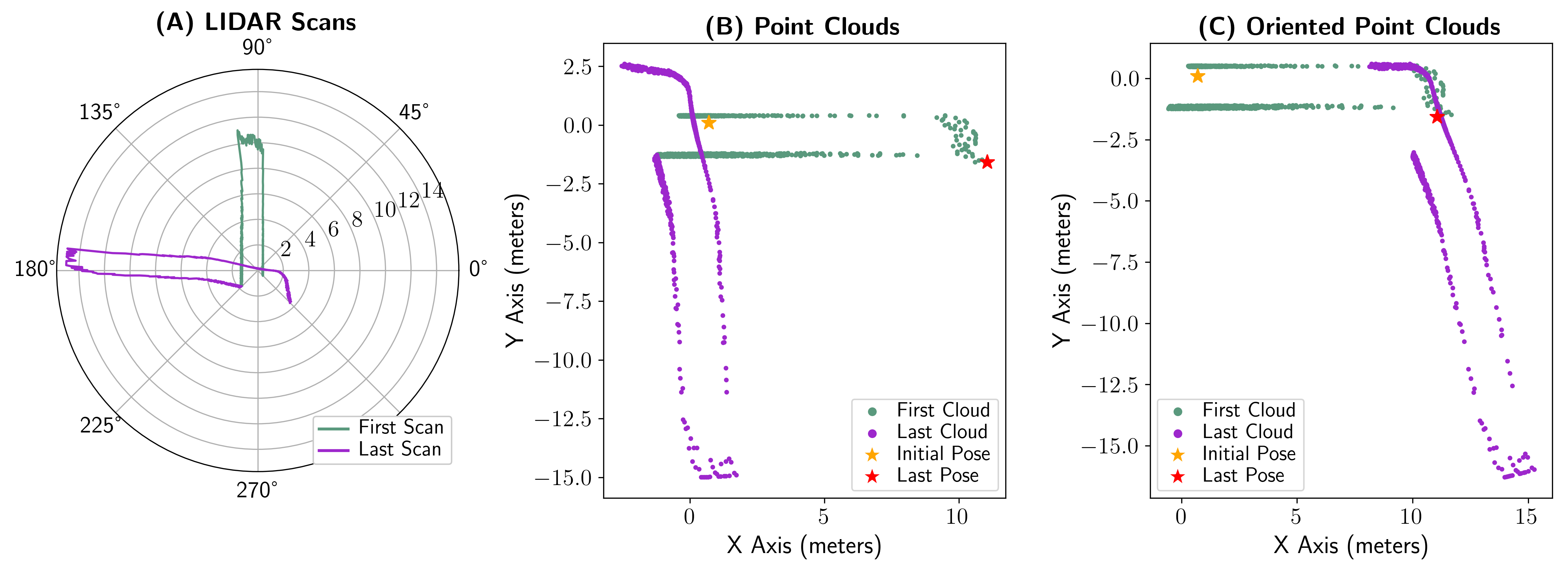}
    \caption{The process of transforming the baseline observation spaces with polar-coordinate LIDAR scans to Cartesian coordinates located and orientated on the global reference frame. (A) The first and last LIDAR scans in polar coordinates. (B) The scans transformed into Cartesian coordinates. (C) The point clouds oriented based on the agent's location and orientation.}
    \label{fig:data-transformation}
\end{figure*}

\textbf{\textit{Racecar Gym (RaceCarGym)}}~\cite{Brunnbauer_racecar_gym}: Inspired by the F1-Tenth driving challenge, \textit{RaceCarGym} tasks agent's with the goal of autonomous driving around flat models of real world racetracks. This environment is developed with PyBullet~\cite{coumans2021} to implement vehicle physics with a differential drive kinematics model, steering on the front wheels, and a ray-casting-based LIDAR sensor. All experiments in this work utilize the default physics parameters. The vehicle is equipped with a 2D LIDAR sensor returning point clouds $\gamma$ with $\gamma_n=1080$ beams across a 273.5$^{\circ}$ field of view. Moreover, the vehicle model also includes pose, velocity, and acceleration sensors with configurable noise. To isolate OGM performance from the variability introduced by SLAM algorithms, we disable all position sensor noise. Additionally, ray-cast measurements are subjected to uniformly distributed noise, constrained within $\pm5\%$ of the true cast length, to simulate realistic sensor imperfections. Controlling the vehicle are two actuators adjusting the steering angle and motor speed. The default observation space is the point cloud $\gamma$ and the corresponding pose and velocity information. Likewise, the action space is a real-valued vector, normalized between -1 and 1, representing the steering angle $\delta$ and motor speed $\tau$.

We create a custom reward function to encourage the agent to maintain a high velocity throughout the tracks whilst minimizing the amount of steering input, the distance to any obstacles, and avoiding collisions. This process results in a final reward value $r$ consisting of four distinct components: $r_{velocity}$, $r_{steering}$, $r_{obstacle}$, and $r_{collision}$. $r_{velocity}$ is designed to encourage the agent to maintain high velocities throughout each timestep such that:
\begin{equation}
    r_{velocity} = \begin{cases}
        \frac{v}{\lambda_0}, & v > \lambda_1 \\
        -1, & \text{else}
    \end{cases}
\end{equation}
where $v$ is the cumulative magnitude of velocities across all dimensions. $\lambda_{0}$ and $\lambda_{1}$ are heuristically chosen to be 3.5 and 0.1 respectively. $r_{steering}$ penalizes very large steering inputs to avoid jerk and is defined as
\begin{equation}
    r_{steering} = \begin{cases}
        0.1, & |\frac{\delta}{\lambda_3}| < \lambda_4 \\
        -\frac{\delta}{\lambda_2}, & else
    \end{cases}
\end{equation}
$\lambda_{2}$, $\lambda_{3}$, and $\lambda_{4}$ are heuristically chosen to be 15, 5, and 0.08, respectively. $r_{obstacle}$ penalizes the agent for being too close to any given hazard and is calculated by subtracting 0.4 from the distance to the closest obstacle. $r_{collision}$ is a fixed penalty of 100 that is applied if the agent hits any obstacle during the course of an episode. Bringing everything together, the final reward value $r$ is calculated as:
\begin{equation}
    r=r_{velocity}+r_{steering}+r_{obstacle}+r_{collision}.
\end{equation}

\textbf{\textit{Environment Wrapping}}: To transform polar coordinate based LIDAR scans to Cartesian space, we create a theta vector $\Theta$ with shape $\gamma_n$ representing the degree of rotation about the unit circle with respect to each ray-cast. For the \textit{MarsExplorer} environment with a 360$^\circ$ field of view and $\gamma_n$ ray-casts, $\theta$ will have a discrete radian step size of $\Delta\Theta=\frac{2\pi}{\gamma_n}$ such that $\Theta=\{0 * \Delta\Theta, 1 * \Delta\Theta, ..., (\gamma_n -1) * \Delta\Theta\}$. Conversely with the \textit{RaceCarGym} having a field of view of 273.5$^{\circ}$, it has a different step size of $\Delta\Theta=\frac{273.5*2\pi}{365*\gamma_n}$ radians.

\begin{algorithm}
    \caption{Polar to Cartesian Coordinates}\label{alg:polar2cartesian}
    \begin{algorithmic}
        \REQUIRE $\gamma_n\gets\text{number of ray-casts}$
        \REQUIRE $\gamma \gets \text{LIDAR scan with } \gamma_n \text{ ray-casts}$
        \REQUIRE $\Theta \gets \text{angle of each ray-cast in radians}$
        \STATE $X\in \mathbb{R}^{N\times 2}$
        \STATE $i \gets 0$
        \WHILE{$i < \gamma_n $}{
            \STATE $X[i, 0] \gets \gamma[i]\sin(\Theta[i])$
            \STATE $X[i, 1] \gets \gamma[i]\cos(\Theta[i])$
            \STATE $i = i + 1$
        }
        \ENDWHILE
        \RETURN $X$
    \end{algorithmic}
\end{algorithm}

\begin{figure*}[ht]
    \centering
    \includegraphics[width=0.9\linewidth]{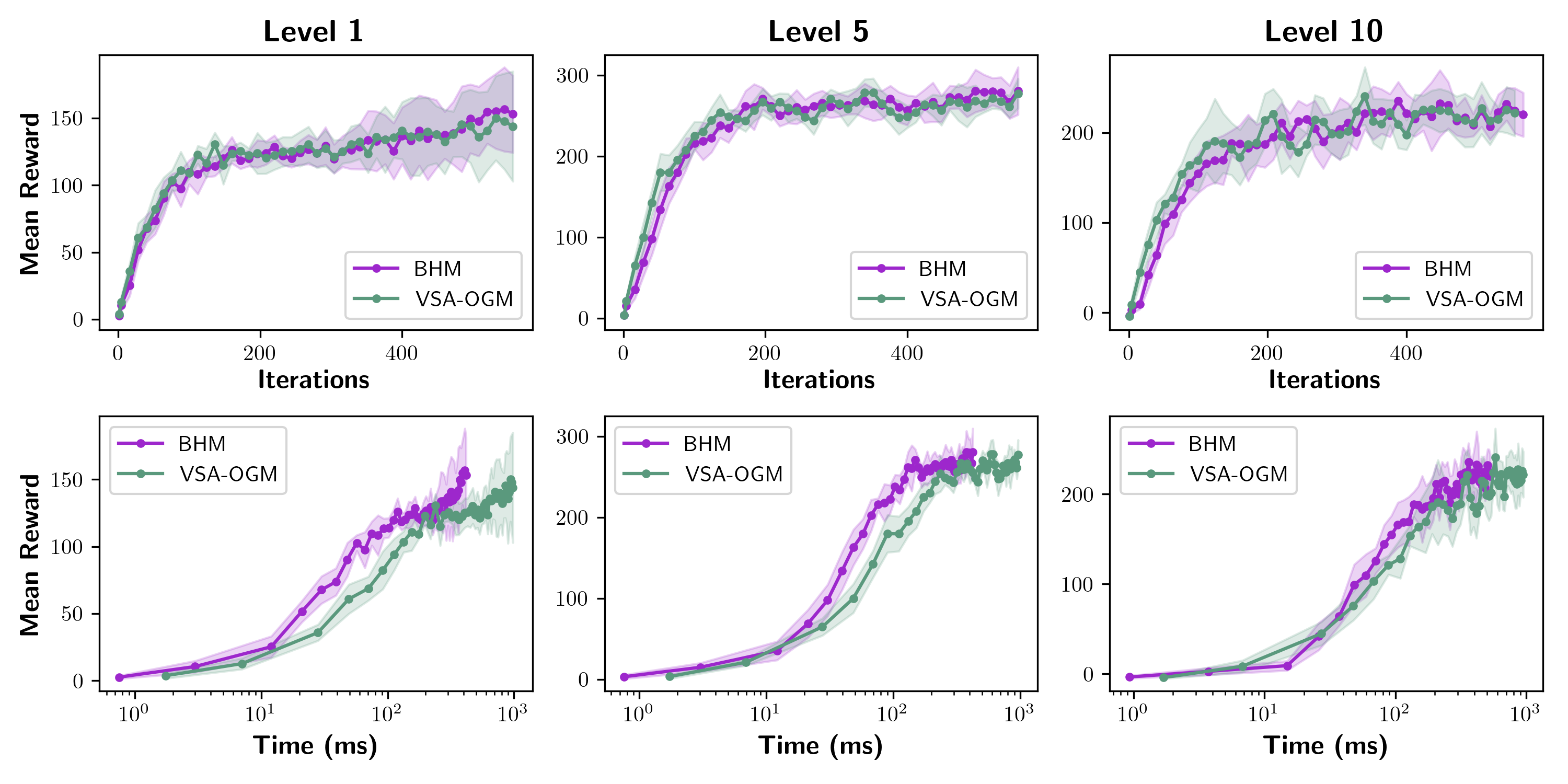}
    \caption{The qualitative results of training a convolutional policy network with Proximal Policy Optimization~\cite{schulman2017proximalpolicyoptimizationalgorithms} for multiple levels in \textit{MarsExplorer}~\cite{Koutras2021MarsExplorer} with VSA-OGM~\cite{snyder2024brain} and BHM~\cite{senanayake2017bayesian}.}
    \label{fig:mars-results}
\end{figure*}

Following Algorithm~\ref{alg:polar2cartesian}, we convert each ray-cast length in $\gamma$ and each ray-cast angle in $\Theta$ to a Cartesian coordinate resulting in a matrix $X\in \mathbb{R}^{\gamma_n \times 2}$. We create a corresponding output matrix $y$ with shape $\gamma_n \times1$ and all values defaulting to 1. To account for ray-casts that do not intersect with an obstacle and return the maximum cast distance $\gamma_{max}$, we employ element-wise conditioning and assignment to set all $y$ values, where indices that match $\gamma[i]=\gamma_{max}$ are set to 0 signifying that the ray-cast corresponds to an empty location.

To avoid under-sampling free-space within the environment, we employ linear interpolation to generate $M$ points along each individual ray-cast. This process generates a matrix $X_{int}$ with $M$ evenly distributed points between 0 and the length of the i-th raycast $\gamma[i]$ resulting in a final matrix shape of $M\gamma_n\times2$. A corresponding target vector $y_{int}$, with shape $M\gamma_n$, is initialized with zeros signifying that all points in $X_{int}$ are classified as empty. We concatenate $X$ and $X_{int}$ along with $y$ and $y_{int}$ resulting in vector-matrices with shape $(M+1)\gamma_n\times2$ and $(M+1)\gamma_n$, respectively. We specify $M=1$ and $M=20$ for all \textit{MarsExplorer} and \textit{RaceCarGym} experiments, respectively. These values were empirically chosen based on the environmental complexity. The last step in our data preprocessing is creating a geometric transformation matrix $T$ to translate and rotate the Cartesian coordinates to the global reference frame using the agent's pose. More information on this process can be found in~\cite{cullen2012matrices}. A visualization of the end-to-end data transformation process is shown in Figure~\ref{fig:data-transformation}.


\section{Results \& Discussion}
\label{sec:results}

\noindent
We evaluate the performance characteristics of BHM~\cite{senanayake2017bayesian} and VSA-OGM~\cite{snyder2024brain} across two distinct RL environments known as \textit{MarsExplorer}~\cite{Koutras2021MarsExplorer} and \textit{RaceCarGym}~\cite{Brunnbauer_racecar_gym}. All RL models are trained using the open-source Stable-Baselines framework~\cite{stable-baselines3} with Open-AI Gym~\cite{1606.01540} and Farama's Gymnasium~\cite{towers2024gymnasiumstandardinterfacereinforcement}. We train models and perform all timing analysis using an NVIDIA DGX compute cluster with 40GB A100 GPUs. It should be noted all experiments are performed on a slice of a single DGX machine with one A100 GPU.

\textbf{Environment Exploration with \textit{MarsExplorer}:} The \textit{MarsExplorer} environment tasks autonomous agents with exploring unknown environments. We create three environments with varying sizes to increase the difficulty of learning the entire environment. These levels are known as Level 1, Level 5, and Level 10 with shapes [20x20], [30x30], and [40x40] respectively. All levels maintain the same reward coefficients as described in Section~\ref{sec:methods}.

Taking inspiration from the default observation space within \textit{MarsExplorer}, we scale the resulting occupancy grid maps from both BHM~\cite{senanayake2017bayesian} and VSA-OGM~\cite{snyder2024brain} to between 0 and 0.5 with the voxel corresponding to the agent's location having a value of 1.0. 
We chose to scale the resulting occupancy values between a subset of the range and set the location value outside of this range. This process avoids the possibility of an intermediate value being confused with the agent's location.
\textit{MarsExplorer} uses a one-to-one mapping between the number of grid cells and the occupancy grid so the resulting OGM's from both methods have the shape $\mathcal{G}$.

\begin{figure}
    \centering
    \includegraphics[width=0.9\linewidth]{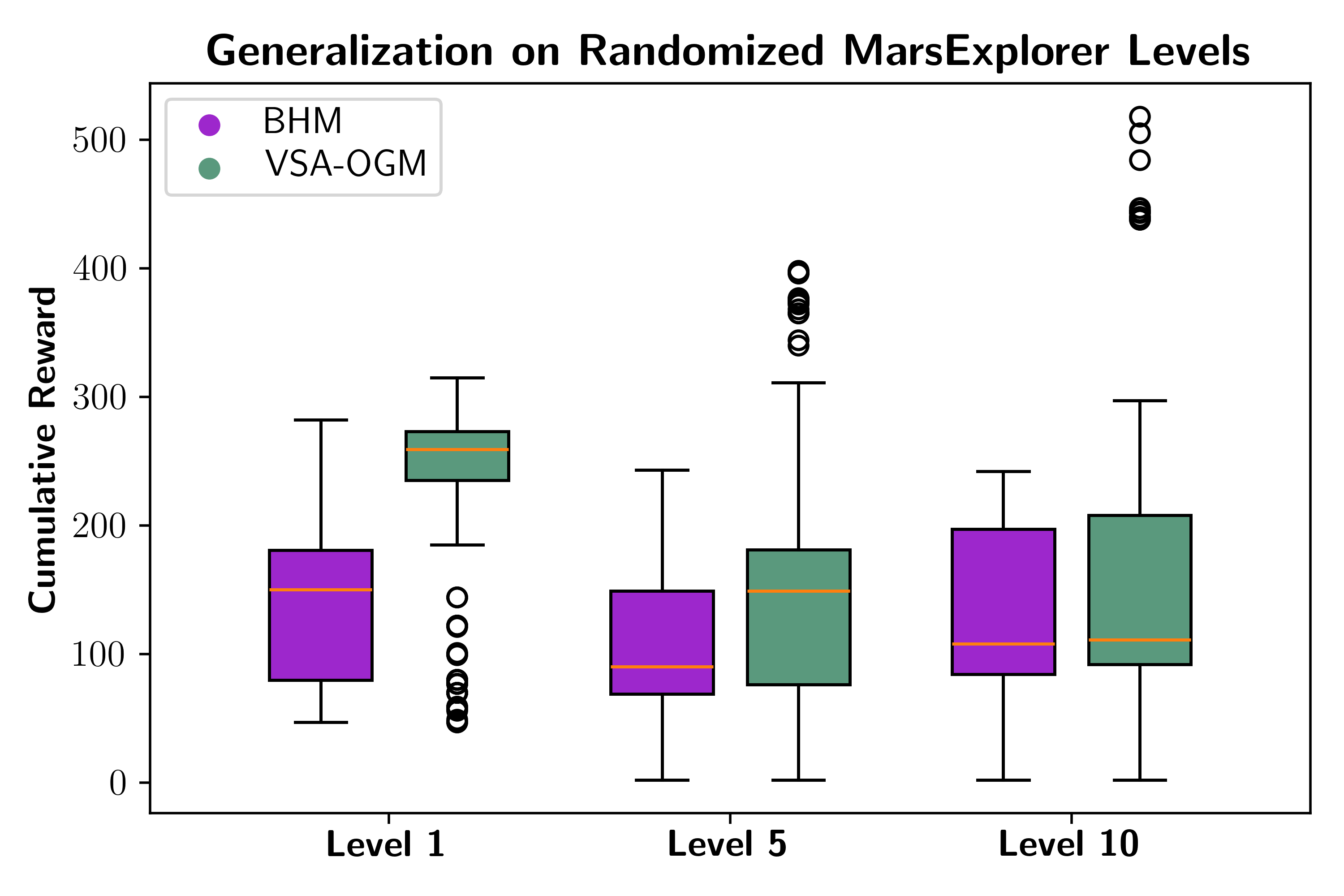}
    \caption{The generalization capabilities of policy networks trained with VSA-OGM~\cite{snyder2024brain} and BHM~\cite{senanayake2017bayesian} on unseen map layouts within the \textit{MarsExplorer} environment~\cite{Koutras2021MarsExplorer}.}
    \label{fig:mars-generalization}
\end{figure}

\begingroup
\renewcommand{\arraystretch}{1.2}
\begin{table}[]
\centering
\caption{Training results with Proximal Policy Optimization~\cite{schulman2017proximalpolicyoptimizationalgorithms} on different levels of the MarsExplorer~\cite{Koutras2021MarsExplorer} environment}
\label{tab:mars-results}
\begin{tabular}{cccc}
\toprule
\textbf{Algorithm} & \textbf{Level} & \textbf{Mean Reward} & \textbf{OGM Latency} \\
\midrule
                                  & 1     &  117.02  & \textcolor{black}{$0.75\pm0.06\text{ms}$}        \\
BHM~\cite{senanayake2017bayesian} & 5     &  230.25  & \textcolor{black}{$0.76\pm0.05\text{ms}$}        \\
                                  & 10    &  180.80   & \textcolor{black}{$0.93\pm0.04\text{ms}$}        \\
\midrule
                           & 1     & 119.10  & \textcolor{black}{$1.76\pm0.32\text{ms}$}        \\
VSA-OGM~\cite{snyder2024brain} & 5     & 232.93  & \textcolor{black}{$1.73\pm0.31\text{ms}$}        \\
                           & 10    & 184.35   & \textcolor{black}{$1.70\pm0.33\text{ms}$}        \\ 
\bottomrule
\end{tabular}
\end{table}
\endgroup

\begin{figure*}[ht]
    \centering
    \includegraphics[width=0.9\linewidth]{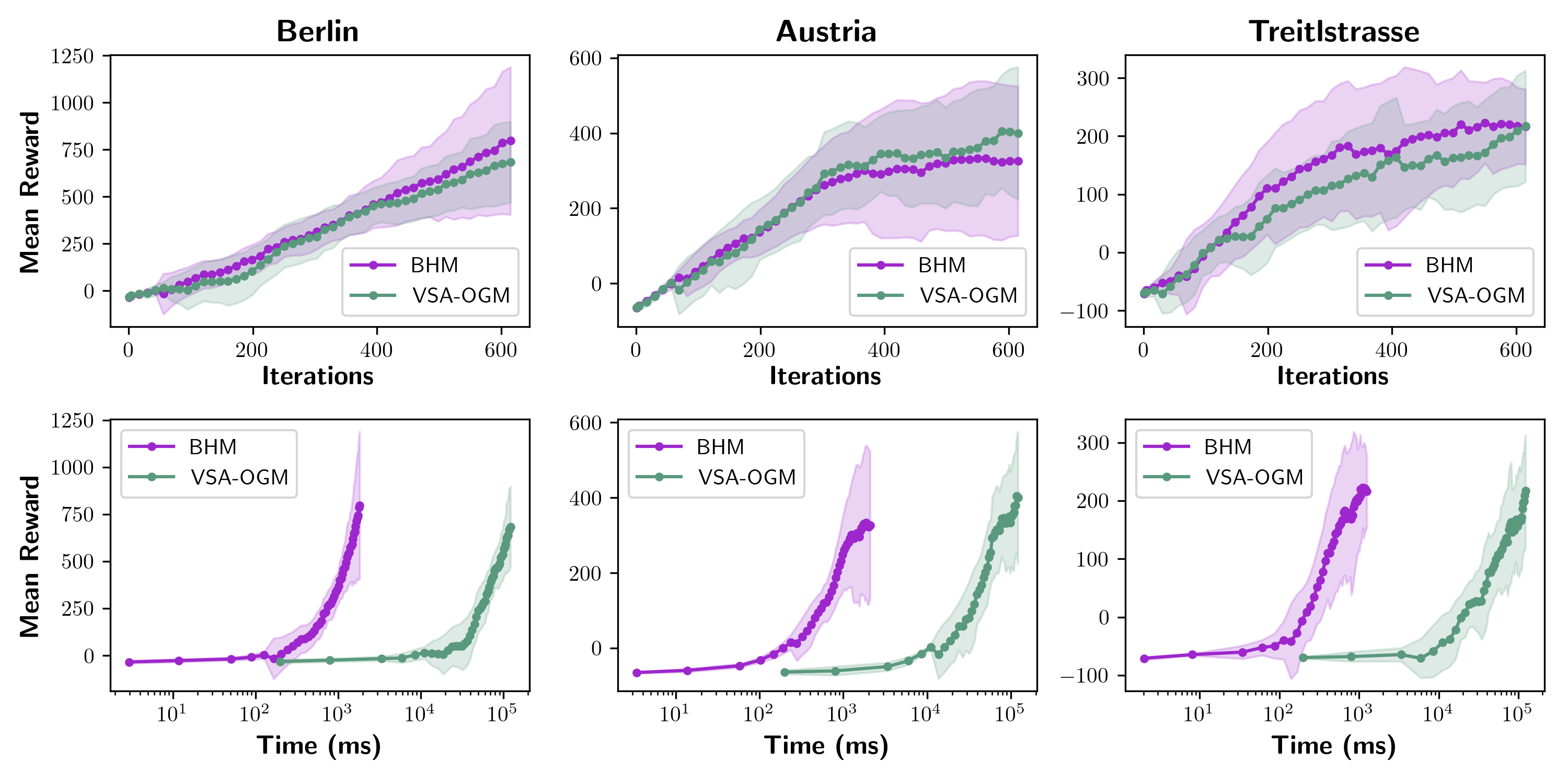}
    \caption{The qualitative results of training a multi-headed policy network with Proximal Policy Optimization~\cite{schulman2017proximalpolicyoptimizationalgorithms} for multiple tracks in \textit{RaceCarGym}~\cite{Brunnbauer_racecar_gym} with VSA-OGM~\cite{snyder2024brain} and BHM~\cite{senanayake2017bayesian}.}
    \label{fig:racar-results}
\end{figure*}

BHM is configured with a bandwidth of 6, while VSA-OGM is configured with a vector dimensionality of 4096, a length scale of 3, and 4 tiles per dimension. In both cases, the length scale and bandwidth parameters control the fidelity and smoothness of the map with respect to map size. The vector dimensionality and number of tiles are unique to VSA-OGM and allow the algorithm to scale based on problem requirements.
The policy network utilizes a convolutional feature extractor with a multi-layer perception (MLP).

The convolutional feature extractor receives the modified OGM and passes it through two convolutional layers. The first layer returns 8 features with a kernel size of 3 and a stride of 2. The second convolutional layer accepts the 8 features and returns 16 features with a kernel size of 2 and a stride of 2. The extracted features are flattened and passed through a linear layer returning 256 features. These features are then passed through a MLP using 4 hidden layers with 64 hidden neurons and ReLU activations. To introduce more variability in the training process, we adjust the decay factor and learning rate with unique values (.95, .99, .999) and (.0003, .000198, .000099), respectively. We train our model with Proximal Policy Optimization (PPO)~\cite{schulman2017proximalpolicyoptimizationalgorithms} and each parameter combination, for 100000 time steps, leading to nine training configurations per OGM algorithm and level combination. The averaged performance of these nine combinations are shown in Table~\ref{tab:mars-results} and Figure~\ref{fig:mars-results}. All timing analysis is performed with PyTorch~\cite{paszke2019pytorchimperativestylehighperformance}, all tensors pre-loaded into GPU memory, and include synchronization between the CPU and GPU.

\begin{figure*}[t]
    \centering
    \includegraphics[width=\linewidth]{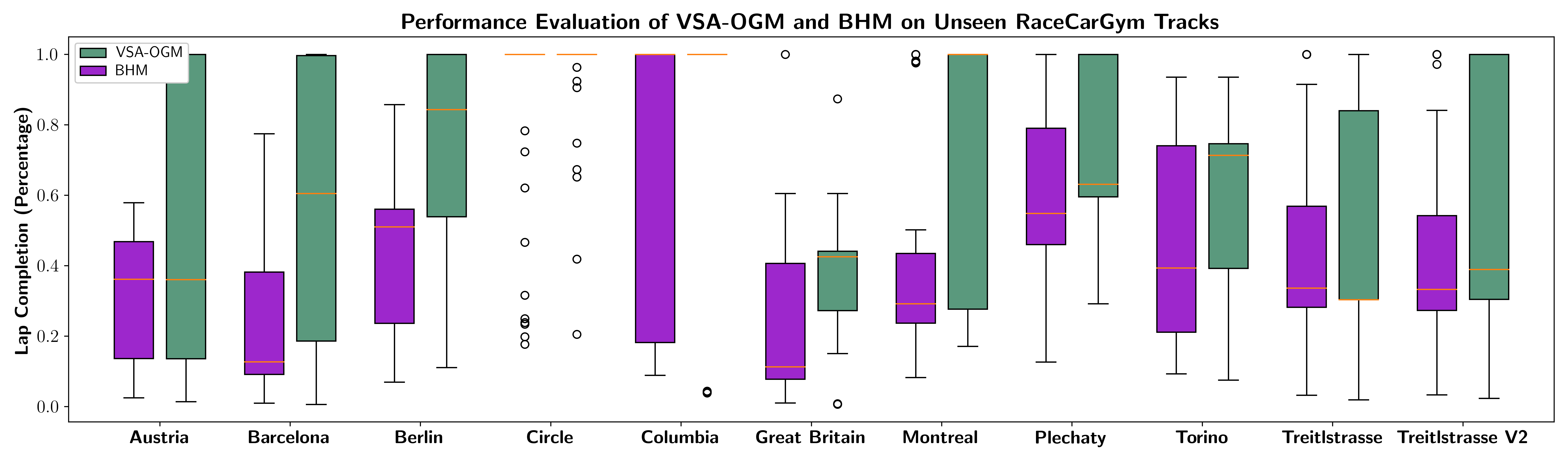}
    \caption{The generalizability of trained multi-headed policy networks when trained and evaluated on multiple maps. Austria, Berlin, and Treitlstrasse were used for training with all remaining maps being used for evaluation. All results are averaged over 5 evaluations.}
    \label{fig:generalization-results}
\end{figure*}

As shown in Table~\ref{tab:mars-results}, VSA-OGM performs slightly better in terms of mean reward at the cost of 2.33x increased latency per time step.
The final size of the learned memory with VSA-OGM was 0.52MB with BHM having a model size of 0.02MB on the largest map (Level 10). This increase in model size is expected as the GPU compatible version of BHM assumes independence between voxels and doesn't maintain the full covariance matrix~\cite{snyder2024brain, senanayake2017bayesian}. 

\textbf{\textit{Generalizability.}} As shown in Figure~\ref{fig:mars-generalization}, we evaluate the generalizability of all trained policy networks on each level with random obstacle layouts. We evaluate each parameter combination across both OGM methods and 25 levels of the same size but with different obstacle layouts for a total of 2250 evaluations. Our results show that policies trained with VSA-OGM exhibit an approximately 52\% increase in generalizability (measured by the percent increase in reward) across all evaluations. This suggests that policy networks trained with VSA-OGM provide policy networks with adequate information to perform the tasks while also not allowing the policy to over fit and suffer on unknown environments. It should be noted that the increased variance with VSA-OGM is expected as PPO is very sensitive to parameter configurations so many of the configurations fail to generalize. Therefore, the increased variance is caused by the high performance ceiling provided by policy networks trained with VSA-OGM.

\begin{figure}
    \centering
    \includegraphics[width=0.75\linewidth]{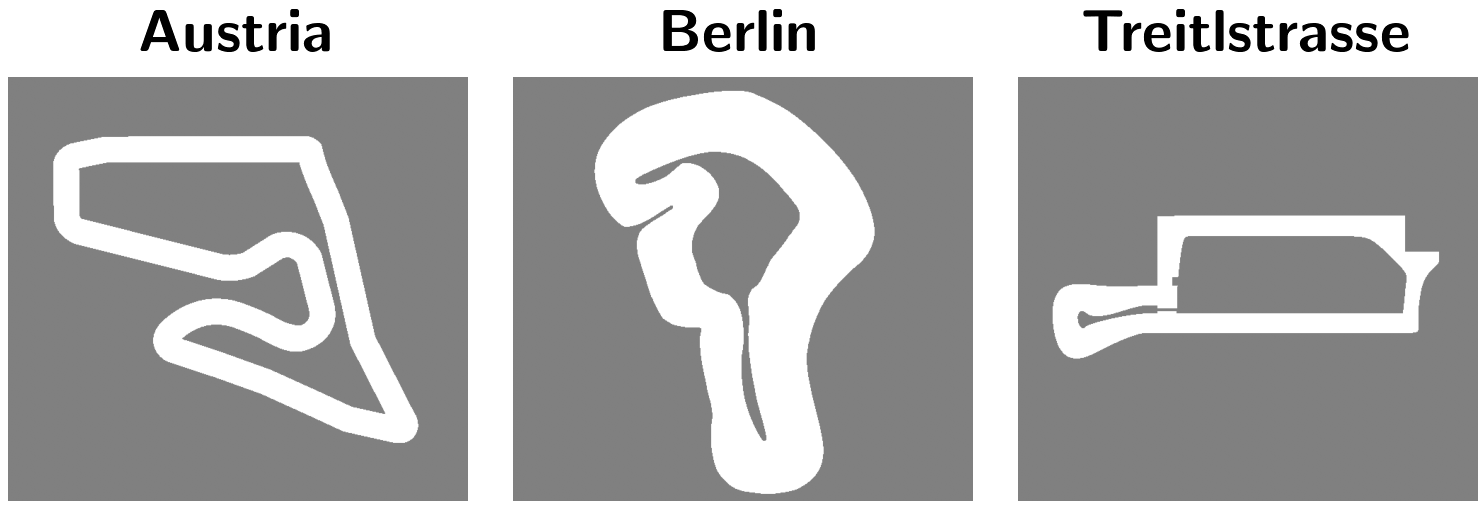}
    \caption{The training maps used within the \textit{RaceCarGym} environment.}
    \label{fig:maps}
    \vspace{-10pt}
\end{figure}

\textbf{Autonomous Driving with \textit{RaceCarGym}:} We evaluate the performance attributes of VSA-OGM for autonomous driving with the \textit{RaceCarGym}~\cite{Brunnbauer_racecar_gym} reinforcement learning environment. We utilized three of the tracks (\textit{Berlin}, \textit{Austria}, and \textit{Treitlstrasse}). \textit{Berlin} is the easiest track with multiple gradual corners with \textit{Austria} having more short radius (closer to $90^\circ$) corners. \textit{Treitlstrasse} is the most difficult with extremely sharp turns through narrow corridors. We train all model configurations individually on each track. A visualization of the three training maps is shown in Figure~\ref{fig:maps}.

\begingroup
\renewcommand{\arraystretch}{1.2}
\begin{table}[]
\centering
\caption{Training results with Proximal Policy Optimization~\cite{schulman2017proximalpolicyoptimizationalgorithms} on different tracks of the RaceCarGym~\cite{Brunnbauer_racecar_gym} environment}
\label{tab:racecar-results}
\begin{tabular}{cccc}
\toprule
\textbf{Algorithm} & \textbf{Track} & \textbf{Mean Reward} & \textbf{OGM Latency} \\
\midrule
                                  & Berlin       & 197.99  & $3.39\pm1.27\text{ms}$       \\
BHM~\cite{senanayake2017bayesian} & Austria      & 334.86  & $2.97\pm0.51\text{ms}$         \\
                                  & Treitlstasse & 121.42  & $2.03\pm0.29\text{ms}$         \\
\midrule
                               & Berlin       & 214.36  & \textcolor{black}{$198.83\pm0.52\text{ms}$}  \\
VSA-OGM~\cite{snyder2024brain} & Austria      & 297.31  & \textcolor{black}{$198.73\pm0.68\text{ms}$}  \\
                               & Treitlstasse & 90.62   & \textcolor{black}{$198.02\pm0.94\text{ms}$}  \\ 
\bottomrule
\end{tabular}
\end{table}
\endgroup

\textit{RaceCarGym}'s default observation space doesn't utilize any form of OGM so we augmented the default observation space to include the OGM as a multiple input observation space. The policy network is a multi-headed MLP receiving LiDAR, pose, velocity, and the OGMs from VSA-OGM~\cite{snyder2024brain} and BHM~\cite{senanayake2017bayesian}. Similar to the \textit{MarsExplorer} experiments, the MLP has 4 hidden layers with 64 hidden neurons using ReLU activations along with the augmented OGM's giving the network greater positional awareness with respect to the map. We limit the resolution of each voxel to a 1 meter by 1 meter area. This is much lower than existing literature on OGM methods~\cite{wilson2022convolutional} because the open-source implementation of BHM runs out of GPU memory when querying the learned model to extract denser occupancy grids.

BHM is configured with a kernel length scale of 6, while VSA-OGM is configured with a vector dimensionality of 10000, a binding length scale of 1, and 8 tiles per dimension. We increase the vector dimensionality and number of tiles per dimensional for all \textit{RaceCarGym} experiments to increase the capacity of VSA-OGM with respect to the increased point cloud densities. In both cases, the length scale parameters control the fidelity of the map with respect to map size.

We utilize the same training process as described for the \textit{MarsExplorer} experiments leading to nine training configurations per OGM algorithm and level combination. Due to the increased complexity of the environment, we increase the number of training steps to 300000. These training results are averaged and form the results presented in Table~\ref{tab:racecar-results} and Figure~\ref{fig:racar-results}.

As shown in Table~\ref{tab:racecar-results}, VSA-OGM stays within 12\% of BHM's mean reward on all tracks with the exception of \textit{Treitlstrasse} at a 33\% drop. However, the mean reward plots suggest that the convergence with VSA-OGM, even on \textit{Treitlstrasse}, was slower than BHM. 
The final memory size of the learned representation with VSA-OGM was 5.12MB with the final BHM method having a model size of approximately 0.01MB depending upon the map size. As previously stated, this increase in parameters is expected as the GPU version of BHM assumes independence between cells and doesn't maintain the full covariance matrix. Compared to the previous \textit{MarsExplorer} experiments, we see that VSA-OGM's latency is approximately 2 orders of magnitude worse than BHM. While the size of the maps is similar between the environments, \textit{RaceCarGym} uses approximately 2 orders of magnitude more points per individual LIDAR scan. This suggests that the encoding operation of higher density point clouds is hindering the latency of VSA-OGM in this experiment.

\textbf{\textit{Generalizability.}} In our final experiment, we perform multi-map training and evaluation where policy networks are sequentially trained across three maps (\textit{Berlin}, \textit{Austria}, and \textit{Treitlstrasse}) for 300000 time steps each. As in all previous experiments, we perform training for each OGM method and all nine parameter combinations. As shown in Figure~\ref{fig:generalization-results}, we evaluate each of these trained models across all training tracks and nine unseen evaluation tracks. These results are averaged across five random seeds and initial conditions. The major takeaway from these results is that the policy networks trained with VSA-OGM present no decrease in model generalizability to unseen environments and even improve generalizability across all evaluation tracks. Therefore, models trained with VSA-OGM maintain similar reward curves during training and present increased generalization in unseen environments at the cost of increased computational complexity.


\section{Conclusion}
\label{sec:conclusion}

\noindent
The efficacy of hyperdimensional OGM techniques as an alternative for traditional OGM methods when leveraged for environmental exploration and autonomous driving remains largely unstudied. In this work, we investigate a hyperdimensional OGM technique, known as VSA-OGM~\cite{snyder2024brain}, and compare it against a well-established, traditional, OGM method known as BHM~\cite{senanayake2017bayesian}.
We evaluate the efficacy of VSA-OGM on two distinct reinforcement learning environments known as \textit{MarsExplorer}~\cite{Koutras2021MarsExplorer} and \textit{RaceCarGym}~\cite{Brunnbauer_racecar_gym}. We perform multiple tests with both environments across a range of scenarios and hyper-parameter combinations.

Our results show that VSA-OGM provides downstream policy networks with sufficient information to learn the environment and effectively perform unknown environment exploration and autonomous driving. We also perform multi-map training and evaluation where VSA-OGM presents approximately 47\% increased generalization performance compared to BHM on \textit{RaceCarGym} and 53\% on \textit{MarsExplorer} at the cost of 2 orders of magnitude increased latency because of increased point cloud densities.


Moving forward, we investigate methods to minimize the memory complexity of VSA-OGM by creating information theoretic methods to minimize adding redundant information to the memory hypervectors. We also investigate methods to minimize the encoding complexity with VSA-OGM as this is the major limitation when point cloud density increases.
We also expand the scope of this exploration to also include model-based reinforcement learning algorithms.

\section*{Acknowledgments}
We acknowledge the technical and financial support of the Automotive Research Center (ARC) in accordance with Cooperative Agreement W56HZV-24-2-0001 U.S. Army DEVCOM Ground Vehicle Systems Center (GVSC) Warren, MI. This project was also supported by resources provided by the Office of Research Computing at George Mason University (URL: https://orc.gmu.edu) and funded in part by grants from the National Science Foundation (Award Number 2018631 and 2319619).

\bibliographystyle{IEEEtran}  
\bibliography{references}

\end{document}